\documentclass[letterpaper, 10 pt, conference]{ieeeconf}  % Comment this line out if you need a4paper

\IEEEoverridecommandlockouts                              % This command is only needed if 
\usepackage{graphicx}
\usepackage{amsmath} % assumes amsmath package installed
\usepackage{subcaption}
\title{Towards Reliable Underwater Diver-Robot Interaction: Gesture Design, Interaction Logic, and Real-World Evaluation}
\author{
Yingqi Liu$^{1,2}$,
Kanzhong Yao$^{2}$,
Zimeng Peng$^{2}$,
Yuanbo Bi$^{2}$,
Anran Li$^{1}$,
Zhe Sun$^{2*}$,
and Xuelong Li$^{2*}$%
\thanks{$^{1}$University of Science and Technology of China, Hefei, China.}%
\thanks{$^{2}$Institute of Artificial Intelligence (TeleAI), China Telecom, Shanghai 200232, China.}%
\thanks{Work done during an internship at TeleAI.}%
\thanks{$^{*}$Corresponding authors: Zhe Sun ({\tt\small sunzhe@nwpu.edu.cn}) and Xuelong Li ({\tt\small xuelong\_li@ieee.org}).}%
}

\usepackage[T1]{fontenc}
\usepackage{cite}
\begin{document}

\maketitle
\thispagestyle{empty}
\pagestyle{empty}

%%%%%%%%%%%%%%%%%%%%%%%%%%%%%%%%%%%%%%%%%%%%%%%%%%%%%%%%%%%%%%%%%%%%%%%%%%%%%%%%
\begin{abstract}
Underwater human--robot interaction requires gesture commands that are both easy for divers to use and reliable for robots to recognize. We investigate these aspects through a closed-loop diver--robot interaction framework integrating a compact seven-gesture vocabulary, lightweight landmark-based recognition, and command-level interaction logic. We evaluate the framework through a user study and underwater robot
experiments in a laboratory tank and a swimming pool. The user study supported the reproducibility of the gestures after brief learning. Recognition analysis further showed that visual similarity was associated with gesture confusion, while intermediate poses during gesture formation introduced temporal ambiguity. Command-level processing mitigated the effects of transient recognition errors on robot execution, reducing unintended triggers and premature task interruptions. These findings show that reliable underwater gesture interaction
depends on human usability, gesture recognizability, and execution reliability in underwater interaction.
\end{abstract}

%%%%%%%%%%%%%%%%%%%%%%%%%%%%%%%%%%%%%%%%%%%%%%%%%%%%%%%%%%%%%%%%%%%%%%%%%%%%%%%%
\section{Introduction}
In underwater human-robot collaborative tasks, divers need to communicate task commands and intentions to robotic platforms. However, communication modalities commonly used in terrestrial environments, such as speech and wireless communication, are severely constrained underwater \cite{hozyn2024advancements,kvasic2024dataset}. 
Hand gestures provide a natural alternative because they are already familiar to divers and can be used without additional communication hardware \cite{kvasic2024dataset,codddowney2025diver}. 
% , by contrast, are a natural means of communication among divers and are therefore well suited for diver–robot interaction \cite{kvasic2024dataset,codddowney2025diver}. 
For task-oriented underwater interaction, a gesture vocabulary must support key command semantics while remaining simple and intuitive. Furthermore, it should minimize physical exertion to accommodate divers who are often burdened with heavy equipment and primary operational tasks \cite{chiarella2018gesture,islam2018dynamic}. 
% In practice, divers may already be occupied with the primary task or other equipment, making it desirable for gesture interaction to impose minimal additional effort.

Achieving reliable machine interpretation of underwater gestures is extremely challenging. Underwater visibility, lighting, color distortion, and viewing conditions can significantly alter the appearance of gestures \cite{hozyn2024advancements,kvasic2024dataset,joshi2025oneshot, sun2026extreme}. 
% Variations in visibility, illumination, and observation conditions can further make visual gesture recognition difficult
More importantly, in continuous interaction, gestures are not merely viewed as isolated static poses. Transitions between gestures may
produce transient or unstable recognition outputs, even when the
completed target pose is eventually recognized correctly
\cite{codddowney2025diver,joshi2025oneshot}. 
If these frame-level predictions are directly mapped to robot commands, brief recognition errors could lead to unintended actions, missed commands, or interruptions of ongoing tasks. 
Therefore, reliable underwater gesture interaction requires not only precise visual recognition but also appropriate temporal and command-level interpretation before recognition outputs are translated into robot behavior \cite{islam2019understanding,codddowney2025diver}. 

Existing studies on underwater human-robot interaction have explored artificial visual languages, natural gesture-based communication, and vision-based recognition and command interpretation~\cite{dudek2007robchat,sattar2007fourier,chiarella2018gesture,islam2018dynamic,codddowney2025diver}. 
Marker-based approaches can provide structured and reliable visual communication, but require divers to present additional physical markers and rely on predefined mappings between visual symbols and robot commands~\cite{dudek2007robchat,sattar2007fourier}.
In contrast, natural-gesture approaches avoid such external markers and have demonstrated their feasibility through user studies, recognition experiments, and robotic tasks~\cite{chiarella2018gesture,islam2018dynamic,tan2021communicative,codddowney2025diver}.
However, the relationship between human usability and machine recognition remains insufficiently characterized. 
% -friendly gesture design and machine recognition reliability has received less explicit analysis.
Gestures that are intuitive and easy for divers to learn and reproduce may still be visually similar to other commands, potentially leading to ambiguity in machine recognition.
Furthermore, even if the final poses are visually distinct, intermediate forms during the transition can lead to recognition ambiguity. These observations suggest that reliable underwater gesture interaction requires gesture vocabularies to be designed not only for human intuitiveness and ease of execution, but also for machine-side discriminability. 
% which leads to systematic recognition confusion. 
% are not necessarily equally distinguishable by a machine recognizer, 
%making the coupling between these two design objectives important for reliable underwater interaction.
\begin{figure}[!t]
    \centering
    \includegraphics[width=\columnwidth]{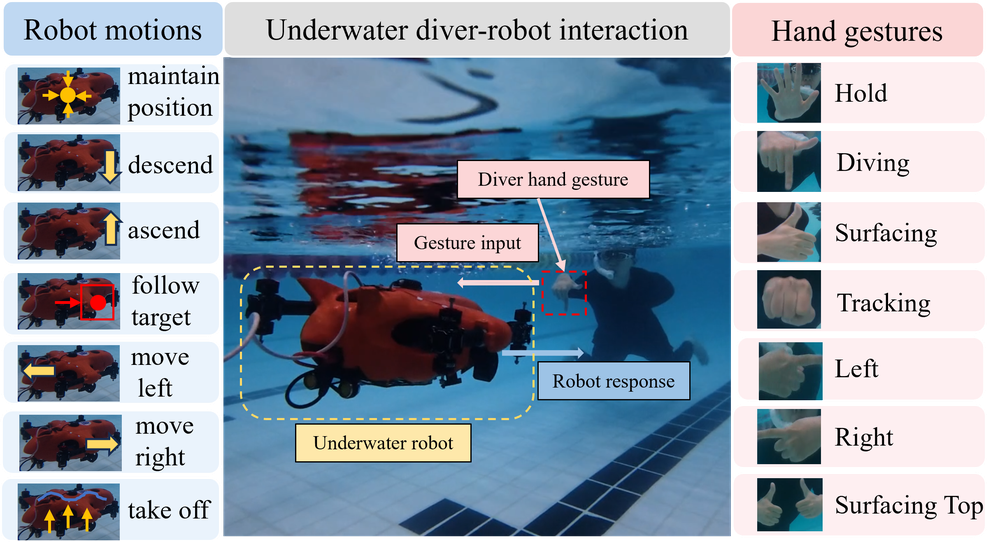}
    \caption{A diver interacting with the underwater robot using hand gestures during swimming-pool experiments.}
    \label{fig:overview}
\end{figure}

In this work, we investigate how underwater gesture interaction can be designed to balance human usability and machine-side recognition reliability. 
% usability for divers and recognition ambiguity for the robot, 
We develop a closed-loop diver-robot interaction framework based on a streamlined command set comprising seven gestures, and evaluate the resulting interaction pipeline from gesture reproduction to underwater recognition and robot command execution. In addition to overall recognition performance, we investigate how visual similarity and gesture transitions lead to recognition ambiguity, and how command-level interaction logic resolves resulting errors during continuous interaction.
The framework is implemented on a physical underwater robot and evaluated in both laboratory-tank and swimming-pool environments (illustrated in Fig.~\ref{fig:overview}). 

% We design a compact seven-gesture vocabulary for underwater diver--robot interaction. We further implement an online diver--robot interaction system as an experimental platform for real-world validation, as illustrated in Fig.~\ref{fig:overview}. The system integrates hand-landmark-based gesture recognition with command-level interaction logic for handling residual recognition errors.
\begin{figure*}[t]
    \centering
    \includegraphics[width=1.0\textwidth]{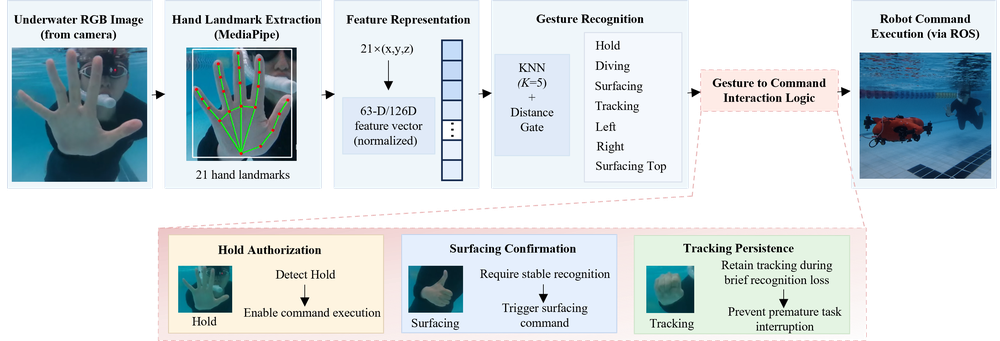}
    \caption{Overview of the proposed underwater diver--robot interaction framework. 
    RGB observations are processed for hand-landmark-based gesture recognition, 
    followed by gesture-to-command interaction logic and robot command execution.}
    \label{fig:system_framework}
\end{figure*}
Our main contributions are summarized as follows:
\begin{itemize}
    \item We develop and deploy an online closed-loop underwater diver-robot interaction framework that integrates lightweight hand-landmark-based gesture recognition with command-level interaction logic, as illustrated in Fig.~\ref{fig:system_framework}, enabling real-time robot control and end-to-end evaluation on a physical underwater platform. 
    % evaluate the proposed gesture vocabulary through a user study, demonstrating its learnability and reproducibility after limited learning.

    \item We quantitatively analyze machine-side recognition ambiguity from both spatial and temporal perspectives. In particular, we relate inter-gesture visual similarity to bidirectional recognition confusion and identify gesture-formation ambiguity caused by intermediate poses.

    \item We validate the proposed gesture interaction scheme on a real underwater robot in closed-loop experiments, and evaluate how interaction-level processing mitigates residual recognition errors during command execution.
\end{itemize}

\section{Related Work}
\subsection{Diver-to-Robot Communication}
Divers can communicate commands and task intentions to underwater robots through dedicated underwater input devices, artificial visual markers, and body-motion-based visual interfaces \cite{birk2022survey,aldhaheri2024review}. Previous systems have employed underwater tablets and instrumented interfaces for direct diver--robot communication \cite{verzijlenberg2010swimming,antillon2023glove}, as well as marker-based visual languages such as RoboChat and FourierTag  \cite{dudek2007robchat,sattar2007fourier}.
Hand gestures and body motions provide a more natural form of interaction. Prior work has used standard SCUBA gestures and custom gesture vocabularies to issue control and task commands to underwater robots \cite{chiarella2018gesture,islam2018dynamic,codddowney2025diver}. Islam \textit{et al.} further enabled divers to issue new commands and reconfigure robot behaviors during underwater missions \cite{islam2018dynamic,islam2019understanding}. Body motion has also been used to convey spatial information, such as the Diver Interest via Pointing (DIP) method and its extension to three-dimensional target indication \cite{edge2023dip,edge2026dip3d}. 
%Other studies have extended gesture-based interaction to broader diver--robot collaboration and multi-robot underwater scenarios \cite{nad2019adriatic,enan2022robotic}.

\subsection{Gesture Recognition and Command Interpretation}

Underwater gesture recognition is challenged by variations in visibility, illumination, and color distortion, which can significantly affect gesture appearance \cite{gomezchavez2021robustness,birk2022survey}. Existing studies have therefore explored a wide range of recognition methods, including conventional visual features, monocular hand-gesture recognition, CNN-based detectors, lightweight embedded models, visual--textual matching, and more recent Transformer-based architectures \cite{martija2020underwater,zahn2020underwater,gomezchavez2021robustness,liu2022ssdlite,zhang2023visualtextual,wang2025dgrformer}. Other work has investigated gesture segmentation, subject-independent recognition, and robustness under degraded underwater visibility \cite{abdi2026underwater}.

Deep visual models generally require sufficiently large and diverse training datasets. Accordingly, several underwater gesture datasets have been developed at considerable scale to support data-driven recognition. The CADDY dataset provides annotated stereo imagery collected across multiple underwater scenarios, while Kvasić \textit{et al.} released a dataset containing more than 30,000 underwater frames and nearly 900 gesture instances recorded with multiple divers under different environments and viewing distances \cite{gomezchavez2019caddy,kvasic2024dataset}. However, collecting and annotating real underwater gesture data remains costly and labor-intensive \cite{kvasic2024dataset}.
To reduce this data burden, recent work has explored zero-shot and one-shot recognition, which aim to recognize new gesture classes with no class-specific training samples or only a single example, respectively \cite{sarma2024zeroshot,joshi2025oneshot}.
Beyond recognition performance itself, several studies have considered how recognition outputs are converted into robot commands. Gomez Chavez \textit{et al.} introduced syntactic validation to reject invalid command sequences \cite{gomezchavez2018robust}, while Islam \textit{et al.} used a finite-state machine and consecutive-frame consistency for gesture-to-instruction mapping \cite{islam2018dynamic}. Codd-Downey and Jenkin further incorporated confirmation at both gesture and gesture-sequence levels before command execution \cite{codddowney2025diver}. 

\section{Methodology}
In this section, we present the design and implementation of our underwater gesture-based human--robot interaction framewor, including gesture recognition  and gesture-to-command interaction logic.
\subsection{Command-Gesture Set Design} 
\begin{figure*}[t]
    \centering
    \includegraphics[width=\textwidth]{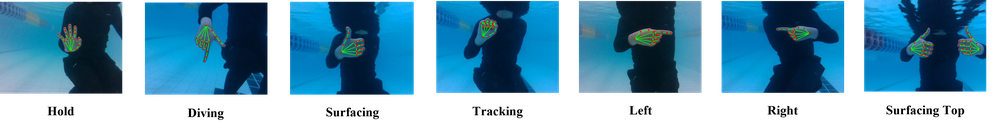}
    \caption{The proposed set of communicative gestures with detected hand landmarks.}
    \label{fig:gesture_set}
\end{figure*}

Based on the practical interaction requirements of underwater robots, we define seven command--gesture mappings using gesture classes labeled: \texttt{Hold}, \texttt{Diving}, \texttt{Surfacing}, \texttt{Tracking}, \texttt{Left}, \texttt{Right}, and \texttt{Surfacing Top}.
For each command, we design a corresponding gesture following principles commonly adopted in underwater communication, favoring gestures that are intuitive, semantically meaningful, and easy to perform
\cite{tan2021communicative,chiarella2018gesture,islam2018dynamic}. Whenever possible, the gesture is chosen to reflect the corresponding robot motion or task semantics, making the intended command easier to interpret and remember. We also consider ease of execution underwater by avoiding large body movements and complex gesture combinations. Except for \texttt{Surfacing Top}, all gestures can be performed with a single hand. The resulting gesture set is illustrated in Fig.~\ref{fig:gesture_set}, and the corresponding command--gesture mappings are summarized in Table~I.
\begin{table}[t]
\caption{Command Semantics and Corresponding Robot Behaviors}
\label{tab:command_set}
\centering
\small
\begin{tabular}{lc}
\hline
\textbf{Command} & \textbf{Robot Behavior} \\
\hline
Hold & Maintain position \\
Diving & Descend \\
Surfacing & Ascend \\
Tracking & Track the target \\
Left$^{*}$ & Move laterally to the left \\
Right$^{*}$ & Move laterally to the right \\
Surfacing Top & Ascend to the surface and take off \\
\hline
\multicolumn{2}{l}{\footnotesize $^{*}$Left/Right are defined from the diver's perspective.}
\end{tabular}
\end{table}
\subsection{Gesture Recognition}
Edge inference \cite{shao2026aiflow} and neural video compression \cite{yuan2026positive} address computation and communication constraints in visual systems.
To support real-time deployment and rapid adaptation of the gesture vocabulary, we adopt a lightweight landmark-based recognition pipeline. The \mbox{MediaPipe} Hand Landmarker is used to extract hand landmarks \cite{zhang2020mediapipehands}, followed by K-nearest neighbors (KNN) classification with a distance-based rejection gate. The reference set contains 700 feature samples per gesture class, including natural variations in hand orientation and viewing angle. New or modified gestures can be incorporated by updating the reference set without retraining. A distance-based rejection threshold is used to suppress uncertain KNN predictions.
\begin{figure}[t]
    \centering
    \includegraphics[width=\columnwidth]{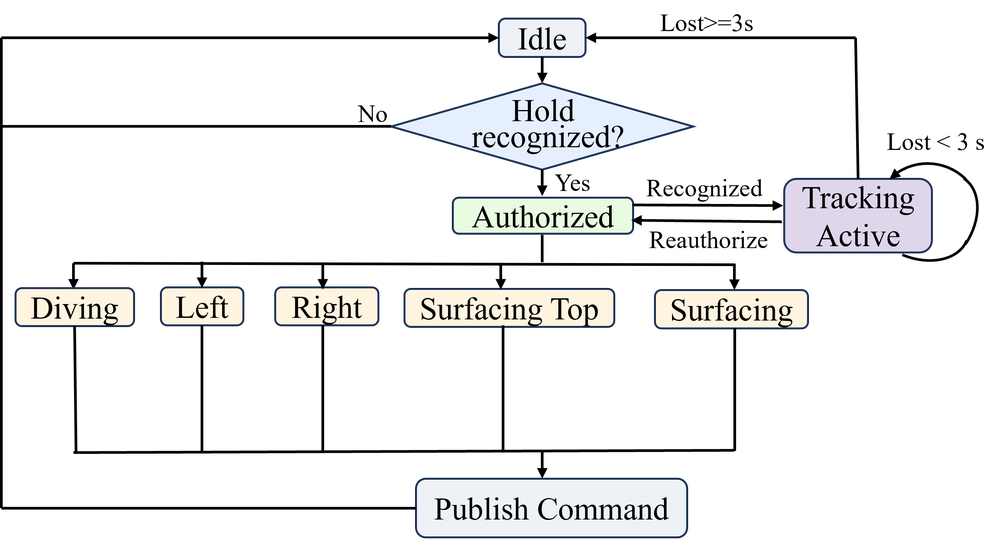}
    \caption{Finite-state machine of the core gesture-to-command interaction
    logic. Hold opens a 5-s authorization window. One-shot task gestures are
    issued within the authorization window, while Tracking persists for up to
    3~s after the latest valid recognition.}
    \label{fig:fsm_logic}
\end{figure}
\subsection{Gesture-to-Command Interaction Logic}
During continuous underwater human--robot interaction, recognition errors and short-term disturbances may produce transient or incorrect outputs \cite{islam2018dynamic,codddowney2025diver}. We therefore introduce gesture-to-command interaction logic between the recognition module and the robot control interface, as illustrated in Fig.~\ref{fig:fsm_logic}.

\textit{Hold-based authorization.}
Hold serves both as a station-keeping command and as an authorization gesture for subsequent motion commands. Once a valid Hold is detected, the robot maintains its position and a 5~s authorization window is opened. The first valid task gesture detected within this window is converted into the corresponding robot command, after which the authorization is cleared. If no valid task gesture is detected within 5~s, the authorization expires and a new Hold is required.

%\textit{Temporal confirmation for Surfacing.}
%To reduce the risk of unintended ascent caused by transient recognition fluctuations, Surfacing requires temporal confirmation before command execution. The command is issued only when Surfacing remains recognized for 0.3~s; otherwise, no Surfacing command is generated.

\textit{Tracking persistence.}
Tracking is a continuous control state and can be disrupted by short-term recognition interruptions, which are common in underwater gesture recognition \cite{zhang2025action,islam2018dynamic}.Once an authorized Tracking command is accepted, the Tracking state is activated immediately and retained for up to 3~s after the latest valid Tracking recognition. Any subsequent Tracking detection within this period resets the timeout; otherwise, Tracking terminates when the timeout expires. Hold has the highest interaction priority and immediately terminates Tracking while opening a new authorization cycle.

\section{Experiments and Results}
We evaluated the proposed gesture set and interaction framework from human and machine perspectives. A user study assessed gesture learnability and reproducibility against two existing underwater gesture vocabularies, followed by laboratory-tank and swimming-pool experiments evaluating gesture recognition and gesture-to-command interaction.
\subsection{Gesture Learnability Evaluation}
To evaluate gesture learnability and reproducibility, we compared six semantically matched gestures from our proposed vocabulary, RoboChatGest, and SGD11: \texttt{Hold}, \texttt{Diving}, \texttt{Surfacing}, \texttt{Tracking}, \texttt{Left}, and \texttt{Right}. Since an easy-to-perform gesture should be readily understood, remembered, and reproduced by users, we quantified this property using the gesture reproduction success rate after limited exposure. Thirty participants were randomly assigned to three groups of ten, with each participant learning only one gesture set. None of the participants had prior diving experience or prior familiarity with the evaluated underwater gesture vocabularies. During the learning phase, each gesture was demonstrated once using a 5-s video, resulting in a total learning time of 30~s. The gesture demonstrations were then removed, and participants were prompted with the task semantics only and asked to reproduce the corresponding gesture within 3~s. All responses were independently evaluated by two annotators.
The gesture reproduction success rate for gesture set $s$ was defined as
\begin{equation}
R_s =
\frac{N_{\mathrm{correct},s}}
{N_{\mathrm{total},s}},
\end{equation}
where $N_{\mathrm{correct},s}$ denotes the number of correctly reproduced gestures and $N_{\mathrm{total},s}$ denotes the total number of trials for gesture set $s$.
The six evaluated gestures from the proposed set achieved a reproduction success rate of 93.3\% (56/60), compared with 85.0\% (51/60) for SGD11 and 65.0\% (39/60) for RoboChatGest. A Kruskal--Wallis test showed a significant difference among the three gesture sets ($H=14.32$, $p<0.001$). Pairwise comparisons with Holm correction further showed that the Proposed set significantly outperformed RoboChatGest ($p=0.0019$), whereas the difference between the proposed gesture set and SGD11 was not statistically significant ($p=0.1222$). These results indicate that the six evaluated gestures can be readily learned and reproduced after brief exposure.

\subsection{Underwater Experimental Setup}
\begin{figure}[t] 
    \centering 

    \begin{minipage}[b]{0.48\columnwidth}
        \centering
        \includegraphics[
            width=\linewidth,
            trim=0 0 0 0,
            clip
        ]{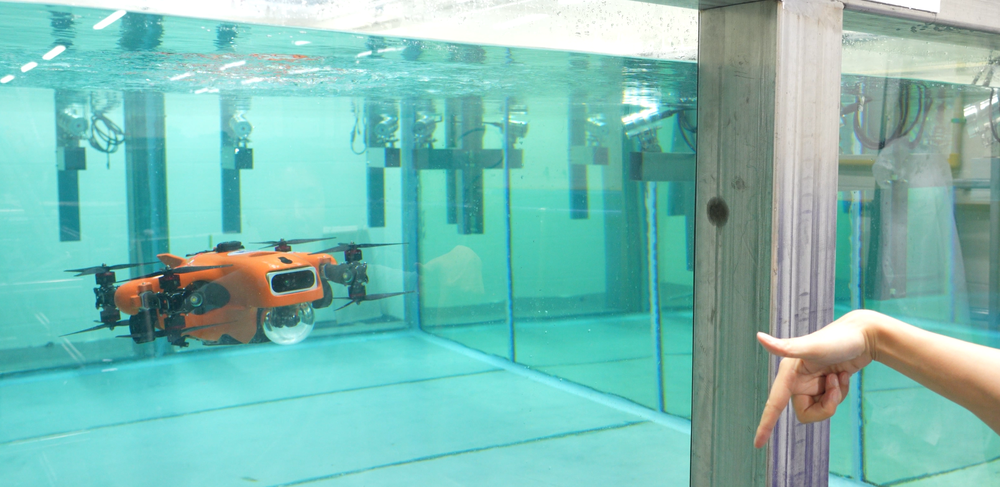}

        \vspace{-1mm}
        {\footnotesize (a)}
    \end{minipage}
    \hfill
    \begin{minipage}[b]{0.48\columnwidth}
        \centering
        \includegraphics[
            width=\linewidth,
            trim=220 110 220 0,
            clip
        ]{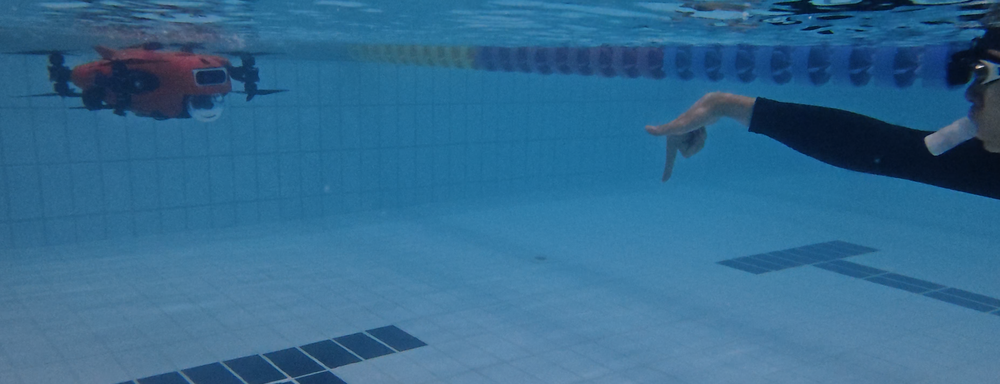}

        \vspace{-1mm}
        {\footnotesize (b)}
    \end{minipage}

    \caption{Experimental settings. (a) Laboratory-tank setting, where the robot operated underwater while gestures were performed in air outside the tank. (b) Swimming-pool setting, where divers interacted with the robot underwater.} 
    \label{fig:exp_setup} 
\end{figure}

We evaluated the proposed diver--robot interaction framework under two experimental settings, as shown in Fig.~\ref{fig:exp_setup}: a laboratory-tank setting and a sports swimming-pool setting. Both settings used the same gesture command set and diver--robot interaction framework. The experimental platform was a 4-DOF hybrid underwater vehicle equipped with an Intel RealSense D435i camera, and the onboard perception and gesture-recognition pipeline was executed on an NVIDIA Jetson Orin NX. During all experiments, the diver remained within the camera field of view so that the hand gestures could be observed by the onboard vision system.
The two settings differed in how the gestures were performed. In the laboratory-tank setting, the robot operated underwater while gestures were performed in air outside the tank, yielding 37 Hold--task interactions (74 gesture inputs). In the sports swimming-pool setting, two divers performed the same gesture set underwater, yielding 56 interactions (112 gesture inputs). Across the two settings, 93 Hold--task interaction sequences were collected in total.
\subsection{Gesture Recognition Performance
}
\begin{figure}[t]
    \centering

    \begin{minipage}[b]{0.48\columnwidth}
        \centering
        \includegraphics[width=\linewidth]{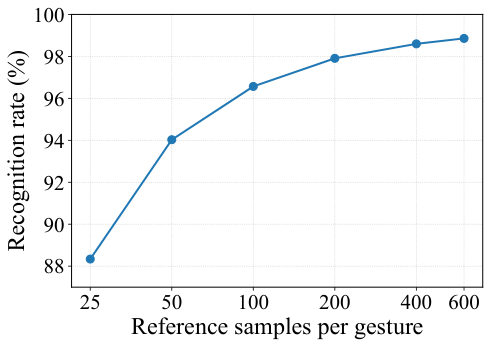}
        \vspace{-1mm}
        {\footnotesize (a)}
    \end{minipage}
    \hfill
    \begin{minipage}[b]{0.48\columnwidth}
        \centering
        \includegraphics[width=\linewidth]{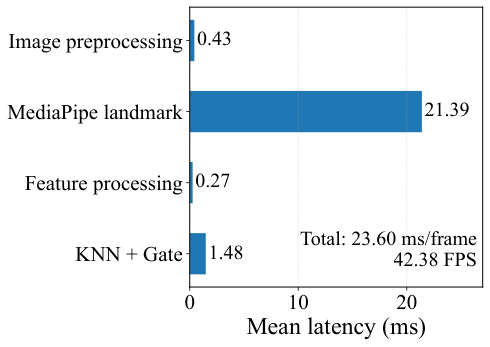}
        \vspace{-1mm}
        {\footnotesize (b)}
    \end{minipage}

    \caption{Efficiency analysis of the landmark-based gesture recognizer: (a) sample efficiency across the seven gestures and (b) runtime breakdown on the NVIDIA Jetson Orin NX.}
    \label{fig:recognizer_efficiency}
\end{figure}

We first assessed the sample and computational efficiency of the
recognizer. In a held-out landmark analysis, accuracy reached 96.57\% and 97.91\% with 100 and 200 reference samples per class, respectively, and improved only marginally thereafter
(Fig.~\ref{fig:recognizer_efficiency}(a)). All 700 reference samples per class were retained in deployment because inference remained lightweight: the
complete pipeline required 23.60~ms per hand-present frame
(42.38~FPS), of which KNN classification with distance-based rejection
required only 1.48~ms (Fig.~\ref{fig:recognizer_efficiency}(b)).

We next evaluated the recognition of complete gesture inputs performed during continuous interaction, independently of the downstream gesture-to-command interaction logic. In the swimming pool, 100 of 112 gesture inputs were correctly recognized; in the laboratory tank, 60 of 74 were correctly recognized. Across both environments, 160 of 186 gesture inputs were correctly classified, corresponding to an overall accuracy of 86.02\%. 
\begin{table}[t]
\caption{Per-class gesture-level recognition performance.}
\label{tab:per_class_recognition}
\centering
\small
\begin{tabular*}{\columnwidth}{@{\extracolsep{\fill}}lcccc@{}}
\hline
\textbf{Gesture} & \textbf{N} & \textbf{P (\%)} &
\textbf{R (\%)} & \textbf{F1 (\%)} \\
\hline
Hold          & 93 & 100.00 & 92.47 & 96.09 \\
Diving        & 14 & 100.00 & 92.86 & 96.30 \\
Surfacing     & 30 & 71.43  & 83.33 & 76.92 \\
Tracking      & 15 & 63.16  & 80.00 & 70.59 \\
Left          & 15 & 82.35  & 93.33 & 87.50 \\
Right         & 10 & 100.00 & 90.00 & 94.74 \\
Surfacing Top & 9  & 100.00 & 11.11 & 20.00 \\
\hline
Macro Avg.    & -- & 88.13 & 77.59 & 77.45 \\
\hline
\end{tabular*}
\end{table}
As summarized in Table~II, the seven gestures achieved an overall recognition accuracy of 86.02\%, while the macro recall and macro F1-score were 77.59\% and 77.45\%, respectively, indicating substantial variation in class-wise recognition performance. In particular, Surfacing Top showed the poorest performance, with a recall of 11.11\% and an F1-score of 20.00\%.

\begin{figure}[t]
    \centering
    \includegraphics[width=\columnwidth]{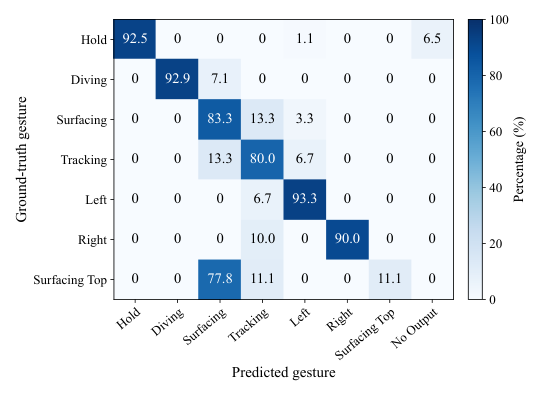}
    \caption{Normalized confusion matrix of gesture recognition.}
    \label{fig:confusion_matrix}
\end{figure}

\begin{figure}[t]
    \centering
    \includegraphics[width=\columnwidth]{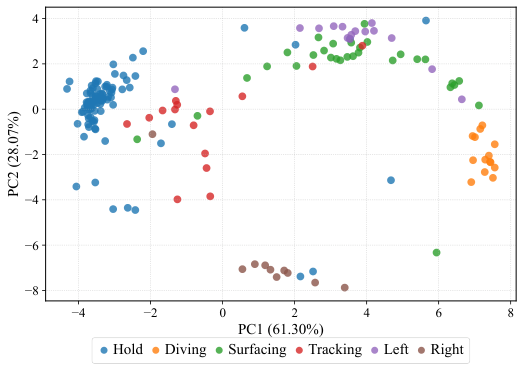}
    \caption{PCA visualization of the feature distributions of the six
    single-hand gesture classes. Each point represents one gesture instance
    using a 63-D hand-landmark feature.}
    \label{fig:pca_features}
\end{figure}

To further understand the sources of recognition errors, we first examined the overall confusion patterns among the seven gesture classes. The normalized confusion matrix in Fig.~\ref{fig:confusion_matrix} shows that a substantial fraction of the errors was concentrated in a small number of gesture pairs, with the dominant confusions occurring between Surfacing Top and Surfacing, and between Surfacing and Tracking. We then analyzed the visual separability of the six single-hand gesture classes to further investigate the relationship between gesture similarity and recognition confusion.
Fig.~\ref{fig:pca_features} shows a PCA projection of the single-hand gesture instances, with the first two principal components explaining 89.4\% of the feature variance and revealing partial overlap among several gesture classes. We therefore quantify visual separability directly in the original 63-D feature space. For gesture class
\(i\), let \(\mathbf{x}_{n}^{(i)}\) denote the feature vector of the \(n\)-th
gesture instance and \(N_i\) the number of instances. We define the class center and within-class dispersion as
\begin{equation}
\boldsymbol{\mu}_i =
\frac{1}{N_i}\sum_{n=1}^{N_i}\mathbf{x}_{n}^{(i)}, \qquad
S_i =
\frac{1}{N_i}\sum_{n=1}^{N_i}
\left\|\mathbf{x}_{n}^{(i)}-\boldsymbol{\mu}_i\right\|_2^2 .
\end{equation}
For a pair of gesture classes \(i\) and \(j\), the squared distance between
their class centers is
\begin{equation}
D_{ij} =
\left\|\boldsymbol{\mu}_i-\boldsymbol{\mu}_j\right\|_2^2 .
\end{equation}
We then define the visual similarity measure as
\begin{equation}
V_{ij} =
\frac{S_i+S_j}{D_{ij}},
\end{equation}
where a larger \(V_{ij}\) indicates greater overlap relative to the
between-class separation and therefore lower visual separability.
To quantify the corresponding recognition confusion, we define the
bidirectional confusion rate between classes \(i\) and \(j\) as
\begin{equation}
C_{ij} =
\frac{n_{i\rightarrow j}+n_{j\rightarrow i}}
     {N_i+N_j},
\end{equation}
where \(n_{i\rightarrow j}\) denotes the number of instances of class \(i\)
misclassified as class \(j\).
We then evaluated the association between \(V_{ij}\) and \(C_{ij}\) across
all single-hand gesture pairs using Spearman's rank correlation coefficient.
A positive rank correlation was observed
(\(\rho=0.6084\), \(p=0.0161\)), indicating that visually more similar gesture
pairs tended to be confused more frequently.

Surfacing Top is a two-hand gesture and therefore has a different feature representation from the remaining single-hand gestures. Among the nine Surfacing Top interactions, seven produced Surfacing as the first recognition output. Examination of the recorded interactions showed that, during some Hold-to-Surfacing Top transitions, one hand first formed the same thumbs-up pose used for Surfacing, followed by the second hand completing the two-hand gesture. This produced a brief intermediate observation visually similar to Surfacing. These observations suggest that, in addition to the visual separability of the final static poses, temporal ambiguity during gesture formation can also contribute to recognition confusion.
\subsection{Evaluation of Gesture-to-Command Interaction Logic
}
To further examine how interaction logic affects the mapping from recognition outputs to robot commands, we replayed the same experimental records offline and progressively introduced different interaction mechanisms. Performance was evaluated over the 93 task interactions by determining whether the intended command was successfully produced in each trial.

We first examine the effect of Hold-based authorization on command triggering. The Direct and Hold-only conditions use the same recognition-output sequences recorded during the experiments and differ only in the control-layer authorization logic. In the Direct condition, a detected task gesture is immediately mapped to its corresponding command. In the Hold-only condition, only the first task command occurring within the 5-s authorization window following a Hold gesture is allowed to pass. We use the Missed Trigger Rate (MTR) to quantify cases in which the intended command fails to be produced: 
\begin{equation}
\mathrm{MTR}
=
\frac{N_{\mathrm{miss}}}{N_{\mathrm{trial}}},
\end{equation}
where $N_{\mathrm{miss}}$ denotes the number of missed task triggers and
$N_{\mathrm{trial}}$ denotes the total number of task interactions. A trial is counted as a missed trigger if no task command matching the ground-truth task intent is published during the interaction.
To quantify erroneous commands triggered between consecutive interactions due to transitional gestures or transient misclassifications, we further define the Inter-trial False Trigger Rate (IFTR) as: 
\begin{equation}
\mathrm{IFTR}
=
\frac{N_{\mathrm{false\text{-}triggered\ intervals}}}
     {N_{\mathrm{inter\text{-}trial\ intervals}}}.
\end{equation}
Because the swimming-pool and laboratory-tank experiments
were conducted as two independent continuous sessions, the
dataset contained 91 inter-trial intervals in total. As summarized in Table~\ref{tab:interaction_logic}, under the
Direct condition, the Missed Trigger Rate was 12.90\% (12/93),
while the Inter-trial False Trigger Rate was 35.16\% (32/91).
With Hold-based authorization, the IFTR decreased to 7.69\%
(7/91), an absolute reduction of 27.47 percentage points.
However, the MTR increased to 30.11\% (28/93). 
Further analysis of the 16 additional missed triggers introduced by Hold-based authorization showed that six resulted from failure to establish authorization after Hold, one from expiration of the authorization window, and the remaining nine from an incorrect task command prematurely consuming the one-shot authorization.
Notably, all nine cases were caused by transient Surfacing recognitions immediately after authorization. Although Hold had been correctly recognized and authorization had been established, a brief Surfacing prediction passed through the control logic before the intended task gesture appeared, preventing the subsequent correct task command from being issued.
\begin{table}[t]
\caption{Comparison of task-triggering performance under different interaction-logic conditions.}
\label{tab:interaction_logic}
\centering
\small
\begin{tabular*}{\columnwidth}{@{\extracolsep{\fill}}lccc@{}}
\hline
\textbf{Condition} &
\textbf{MTR} &
\textbf{IFTR} &
\shortstack[c]{\textbf{Incorrect}\\\textbf{Output}} \\
\hline
Direct
& 12.90\%
& 35.16\%
& -- \\

Hold-only
& 30.11\%
& 7.69\%
& 21.51\% \\

Hold + Surfacing confirmation
& 22.58\%
& --
& 11.83\% \\
\hline
\end{tabular*}
\end{table}
Residual false triggers were still observed in 7 of the 91 inter-trial intervals after introducing Hold-based authorization. Inspection of these cases showed that these residual errors were mainly associated with unintended Hold recognitions during gesture transitions. Because a Hold recognition reopens the task-command pathway, misclassification of the authorization gesture has a larger downstream effect than errors in ordinary task gestures. Overall, Hold-based authorization substantially reduces false triggers outside the authorized interaction window, but introduces additional missed triggers and remains sensitive to false recognition of the authorization gesture itself. This result suggests that an authorization gesture should be not only easy to perform, but also visually distinguishable and reliably recognized, providing an additional design criterion for the gesture set.
\begin{figure}[t]
    \centering
    \includegraphics[width=\columnwidth]{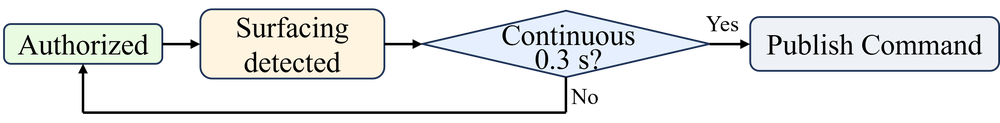}
    \caption{Temporal confirmation logic for the Surfacing command. After
    Surfacing is detected within the authorization window, the command is
    issued only if the gesture is continuously recognized for
    $T_s=0.3$~s; otherwise, the authorized state is resumed.}
    \label{fig:surfacing_confirmation}
\end{figure}

To address the transient Surfacing recognitions observed within the
authorization window, we introduced a temporal confirmation mechanism
for Surfacing while keeping all other control logic unchanged, as
illustrated in Fig.~\ref{fig:surfacing_confirmation}. A Surfacing
command was issued only when the recognition persisted for a confirmation
interval of $T_s=0.3$~s.
With temporal confirmation, the overall Missed Trigger Rate decreased from 30.11\% (28/93) to 22.58\% (21/93), while the proportion of trials containing an incorrect task output decreased from 21.51\% (20/93) to 11.83\% (11/93).
Further examination of the nine transient Surfacing errors showed that temporal confirmation filtered all nine erroneous predictions and allowed the subsequent intended tasks to proceed. The mechanism therefore directly mitigated the dominant failure mode under the Hold-only condition, in which an incorrect task prediction prematurely consumed the one-shot authorization.
This improvement came with a corresponding cost. Of the 30 ground-truth Surfacing trials, 24 were successfully triggered under the Hold-only condition, compared with 22 after temporal confirmation was introduced. Two valid Surfacing gestures failed to satisfy the confirmation requirement because their recognized duration was too short. Among the 22 Surfacing trials successfully triggered under both conditions, temporal confirmation introduced an average additional response delay of approximately 0.392 s.
Overall, Surfacing confirmation filtered transient errors and reduced missed task triggers, but added latency and could reject brief valid gestures.
\begin{figure}[t]
    \centering
    \includegraphics[width=\columnwidth]{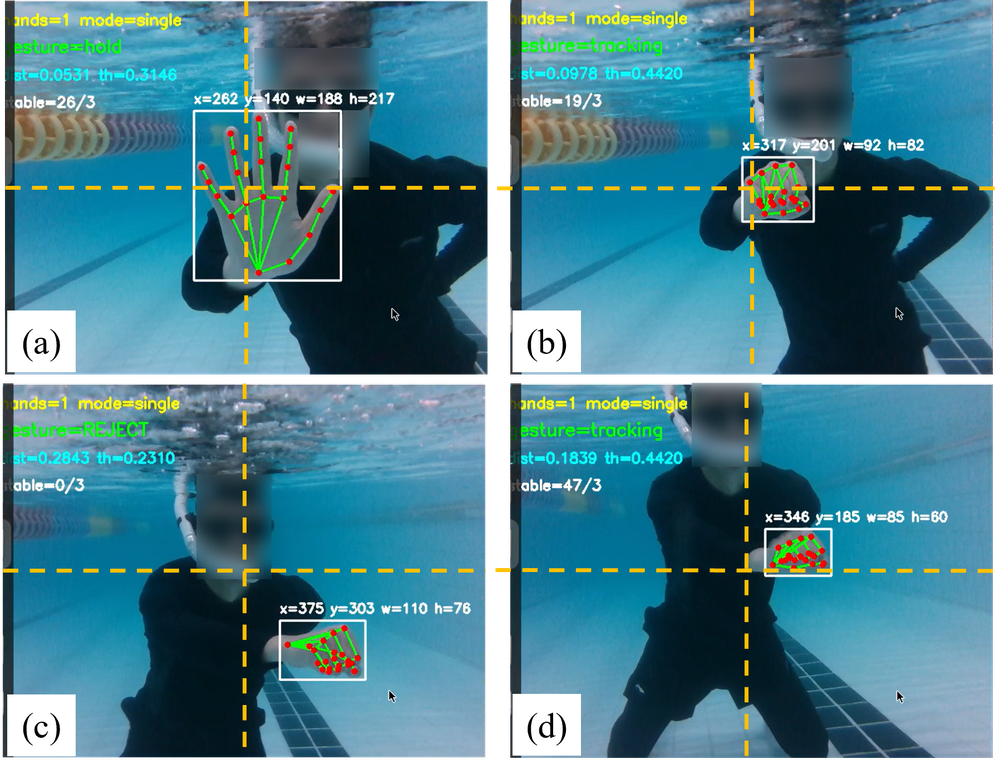}
    \caption{A representative Tracking interaction on the physical underwater robot platform:
    (a) Hold-based authorization;
    (b) the Tracking gesture is recognized and the robot enters the Tracking state;
    (c) a temporary REJECT output occurs while the Tracking state is maintained;
    (d) Tracking is recognized again and the target moves back toward the image center.}
    \label{fig:tracking_demo}
\end{figure}

\begin{figure}[t]
    \centering

    \begin{minipage}[b]{0.48\columnwidth}
        \centering
        \includegraphics[width=\linewidth]{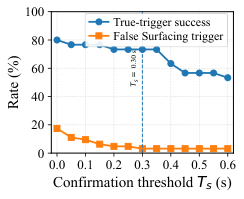}
        \vspace{-1mm}
        {\footnotesize (a)}
    \end{minipage}
    \hfill
    \begin{minipage}[b]{0.48\columnwidth}
        \centering
        \includegraphics[width=\linewidth]{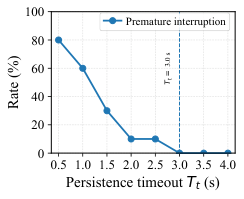}
        \vspace{-1mm}
        {\footnotesize (b)}
    \end{minipage}

    \vspace{-1mm}
    \caption{Sensitivity analysis of the temporal parameters in the gesture-to-command interaction logic: (a) Surfacing confirmation threshold $T_s$ and (b) Tracking persistence timeout $T_t$.}
    \label{fig:temporal_sensitivity}
\end{figure}
Continuous tasks are also susceptible to state interruptions caused by short-term recognition dropouts. Unlike one-shot task commands, Tracking must remain active over a period of time. We therefore compared Tracking continuity with no state persistence and with a 3-s Tracking persistence mechanism. Among the 93 task interactions, we selected 10 Tracking trials that were properly terminated by Hold for analysis.
Among these interactions, 8 would have been prematurely interrupted without state persistence, whereas all 10 remained active until the next valid Tracking recognition or the terminating Hold command under the adopted 3-s persistence.
These results show that Tracking recognition can be discontinuous during real-world underwater interaction, and that short-term state persistence can prevent transient recognition dropouts from directly interrupting a continuous task, thereby improving the continuity and smoothness of the Tracking interaction.
A representative interaction is shown in Fig.~\ref{fig:tracking_demo}. After Tracking is activated, the recognition output briefly changes to REJECT during the diver's hand movement. The persistence mechanism maintains the Tracking state during this interruption, allowing Tracking to continue when Tracking is subsequently recognized again. During this sequence, the target-center error increases from approximately \(43.0\) px to \(149.3\) px during the recognition interruption and then decreases to \(72.9\) px.

To further assess the robustness of the adopted temporal settings, we performed a post-hoc sensitivity analysis by replaying the recorded pre-command recognition streams while varying one temporal parameter at a time and keeping the remaining interaction logic unchanged. For Surfacing, the true-trigger success rate was computed over the 30 Surfacing trials, whereas the false-trigger rate was computed over the 63 non-Surfacing trials. As shown in Fig.~\ref{fig:temporal_sensitivity}(a), increasing the Surfacing confirmation threshold $T_s$ reduced the false-trigger rate from 17.46\% at $T_s=0$ to 3.17\% at $T_s=0.30$~s. Increasing $T_s$ beyond 0.30~s produced no further reduction in false triggers, while the true-trigger success rate decreased from 73.33\% to 63.33\% at $T_s=0.40$~s and continued to decline for larger thresholds. For Tracking, Fig.~\ref{fig:temporal_sensitivity}(b) shows that the premature interruption rate decreased from 80\% at $T_t=0.5$~s to 0\% at $T_t=3.0$~s, with no further improvement for longer timeouts. These results indicate that the adopted $T_s=0.30$~s and $T_t=3.0$~s lie in reasonable operating regions for suppressing transient false triggers and maintaining continuous task execution.
\subsection{End-to-End Underwater Interaction
}
\begin{figure*}[t]
    \centering
    \includegraphics[width=\textwidth]{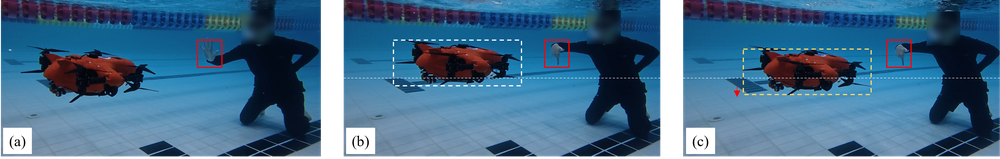}
    \caption{Representative end-to-end underwater diver--robot interaction:
    (a) Hold authorization;
    (b) Diving command;
    (c) robot descent in response to the command.}
    \label{fig:diving_demo}
\end{figure*}

Under the complete interaction logic, we further evaluated the end-to-end performance of the framework from diver gesture input to robot task execution. Across the two experimental settings, the overall command accuracy was 88.17\%. The command accuracies in the swimming pool and laboratory tank were 91.07\% and 83.78\%, respectively.
We further treated each Hold–task sequence as one complete task interaction. Among the 93 task interactions, the robot successfully executed 72 tasks, corresponding to an overall task-execution success rate of 77.42\%. 
To further illustrate the end-to-end interaction process, Fig.~\ref{fig:diving_demo}
shows a representative third-view interaction example from the swimming-pool experiments.
The sequence shows a complete command execution cycle. The diver first performs the Hold gesture to establish authorization, and then issues the Diving gesture. The underwater robot subsequently begins the corresponding descending motion, providing a visual example of successful gesture-to-action execution on the physical platform.

\section{Conclusion}
This work evaluated a closed-loop underwater diver--robot interaction framework from gesture reproduction to recognition and command execution. The results showed that gestures readily reproduced by users can still be ambiguous to a machine because of visual similarity and intermediate poses during gesture formation. Interaction-level processing mitigated the effects of these ambiguities on robot behavior, but involved trade-offs among false triggers, missed commands, and response latency.

These findings suggest that reliable underwater gesture interaction
should be designed and evaluated as an integrated human--machine process. Human usability, machine-side recognition, and robot command execution must be considered together: ease of gesture reproduction or recognition accuracy alone does not establish overall interaction reliability. Broader evaluation across divers and underwater conditions remains necessary.


%%%%%%%%%%%%%%%%%%%%%%%%%%%%%%%%%%%%%%%%%%%%%%%%%%%%%%%%%%%%%%%%%%%%%%%%%%%%%%%%

%%%%%%%%%%%%%%%%%%%%%%%%%%%%%%%%%%%%%%%%%%%%%%%%%%%%%%%%%%%%%%%%%%%%%%%%%%%%%%%%

\bibliographystyle{IEEEtran}
\bibliography{references}

\end{document}